\documentclass[anonymous=false,acmsmall,screen,nonacm]{acmart}
\AtBeginDocument{%
  \providecommand\BibTeX{{%
    \normalfont B\kern-0.5em{\scshape i\kern-0.25em b}\kern-0.8em\TeX}}}

\copyrightyear{2024}
\acmYear{2024}
\setcopyright{rightsretained}
\acmConference[CHI EA '24]{Extended Abstracts of the CHI Conference on Human Factors in Computing Systems}{May 11--16, 2024}{Honolulu, HI, USA}
\acmBooktitle{Extended Abstracts of the CHI Conference on Human Factors in Computing Systems (CHI EA '24), May 11--16, 2024, Honolulu, HI, USA}
\acmDOI{10.1145/3613905.3651044}
\acmISBN{979-8-4007-0331-7/24/05}

\usepackage{soul}

\begin{document}

\title{Field Notes on Deploying Research Robots in Public Spaces}

\author{Fanjun Bu}
\affiliation{%
  \institution{Cornell University, Cornell Tech}
  \city{New York}
  \country{USA}}
\email{fb266@cornell.edu}

\author{Alexandra Bremers}
\email{awb227@cornell.edu}
\affiliation{%
  \institution{Cornell Tech}
  \city{New York}
    \state{New York}
  \country{USA}
}

\author{Mark Colley}
\email{mark.colley@uni-ulm.de}
\orcid{0000-0001-5207-5029}
\affiliation{%
  \institution{Ulm University}
  \city{Ulm}
  \country{Germany}
}
\affiliation{%
  \institution{Cornell Tech}
  \streetaddress{2 W Loop Road}
  \city{New York}
  \country{U.S.}
}

\author{Wendy Ju}
\affiliation{%
  \institution{Cornell University, Cornell Tech}
  \city{New York}
  \country{USA}}
\email{wendyju@cornell.edu}

\renewcommand{\shortauthors}{Bu, et al.}

\begin{abstract}
Human-robot interaction requires to be studied in the wild. In the summers of 2022 and 2023, we deployed two trash barrel service robots through the wizard-of-oz protocol in public spaces to study human-robot interactions in urban settings. We deployed the robots at two different public plazas in downtown Manhattan and Brooklyn for a collective of 20 hours of field time. To date, relatively few long-term human-robot interaction studies have been conducted in shared public spaces. To support researchers aiming to fill this gap, we would like to share some of our insights and learned lessons that would benefit both researchers and practitioners on how to deploy robots in public spaces. We share best practices and lessons learned with the HRI research community to encourage more in-the-wild research of robots in public spaces and call for the community to share their lessons learned to a GitHub repository.
\end{abstract}



\begin{teaserfigure}
\includegraphics[width=\textwidth]{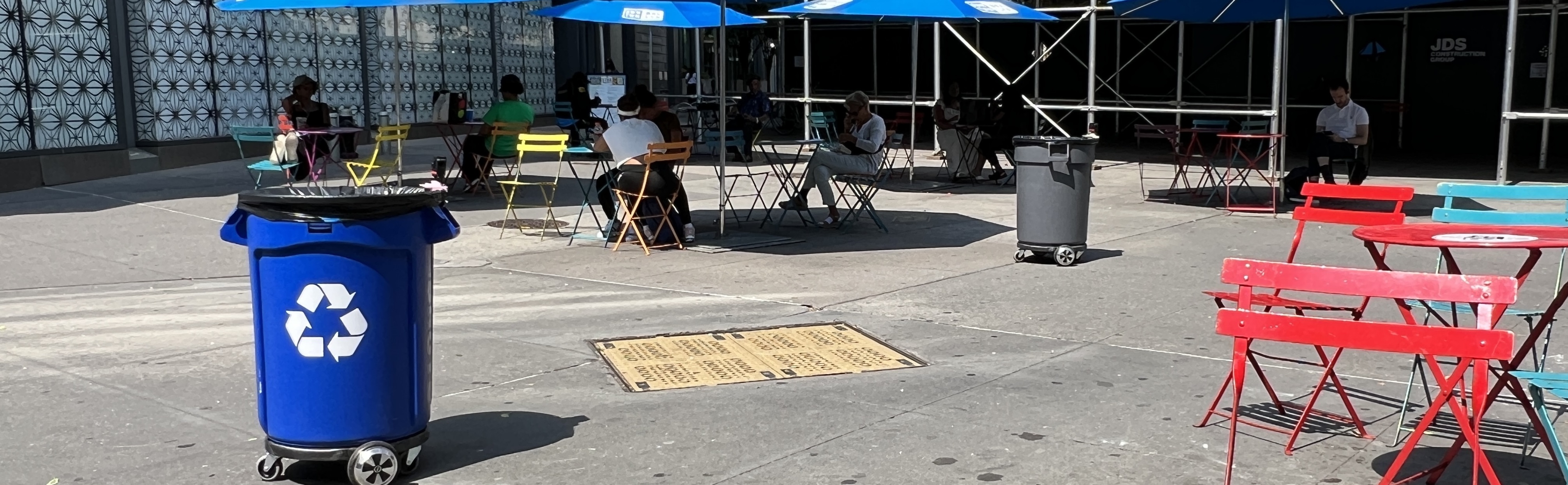}
\caption{Trash barrel robots deployed in a local neighborhood.}
\Description{Two trash barrel robots, one gray for landfill and one blue for recycling, roam around in a public plaza. The public plaza features public seating with colorful chairs and tables under sun umbrellas.}
\label{fig:teaser}
\end{teaserfigure}

\maketitle

\section{Introduction}

Historically, much of the earlier work in HRI research has been based on controlled laboratory experiments. To expand the methodological base of user testing for HRI with in-the-wild approaches, more recent work has looked into adopting methods from other disciplines -- such as ethnography, which offers great methods of qualitative observational studies -- and expanding the use cases of these methods to the field of robotics~\cite{blond2019studying}. 

The 19th Annual ACM/IEEE International Conference on Human-Robot Interaction (HRI), or HRI 2024, the flagship conference in HRI, has been themed "HRI in the real world" to "bring HRI out of the lab and into everyday life"~\cite{hri2024}. As more robots enter peoples' daily lives, it has been increasingly recognized that robots must be piloted outside the lab space~\cite{schneider2022human,svenstrup2009pilot}. Consequently, interaction designs with robots risk being underdeveloped and biased to the limited testing capabilities of lab or virtual reality environments~\cite{WijenTowards}, all the while affecting numerous members of the public in their everyday lives. 

Failure to conduct a thorough evaluation of robots could result in the formulation of policies related to robotics in public spaces occurring only after extensive commercial development and widespread robot deployment. This delay could potentially heighten the risks for individuals in the public domain. Several cities, including Pittsburgh, Miami-Dade County, Detroit, and San Jose, have initiated pilot studies involving personal delivery robots in public areas to facilitate the timely development of informed policies~\cite{PDDP,UrbanismNext}.

However, despite increasing awareness of the need for in-the-wild studies with robotic deployments, barriers remain to deploying research robots in public spaces. For one, as is commonly the case with qualitative studies, each deployment will have unique challenges and factors specific to the type of robot, location, and many other factors. In-person public robotic deployment can also involve significant investment in terms of time and resources, which limits the capability of individual groups to tackle the issues involved. Thus, we are calling for the robotics community to take a stance of openly sharing information and lessons learned so that this grand challenge can ultimately be tackled.

To this end, we want to contribute to a shared knowledge base on HRI-in-everyday-life by sharing our insights and practices from our field deployments of physical service robots in urban settings. We describe our setup and protocol, along with lessons learned and unexpected challenges during deployment. Through this contribution, we offer practical insights to encourage increased robotic deployments during the research phase of robotic interaction design.

To further enable the sharing of this knowledge and engage community participation, we provide a public space in the form of a public GitHub repository, available at \href{https://github.com/FAR-Lab/Roboticists-Field-Insights-and-Guide}{https://github.com/FAR-Lab/Roboticists-Field-Insights-and-Guide}. 

\section{Study Overview}
In our field study, we deployed two trash barrel robots in public plazas in New York City. To date, we operated the robots in two unique neighborhoods, one being a touristy plaza and the other being a local plaza for residents to socialize. The robots were deployed in the early afternoons after the lunch period when the pedestrian traffic at both plazas peaked. The robots were deployed through Wizard-of-Oz; two hidden researchers teleoperated the robots onsite, while plaza users assumed the robots acted autonomously \cite{JD_WoZ,WoZ}. Another onsite researcher interviewed people after they interacted with the robots.  

\section{Guidelines on Deployment}
\subsection{Consent}
Since the study was conducted in the USA, we cannot guarantee that our consent process will generalize to other countries. \nobreak Obtaining consent in public spaces can be tricky. On the one hand, we prefer not to prime participants with flyers and posters, which may either deter people from coming to the plaza or attract people who come solely for the robots; on the other hand, we must inform the participants and obtain consent on the usage of their data.

In our protocol, we adjusted elements of informed consent based on the determination that the research posed minimal risk to the participants and that altering the consent process was necessary for practical reasons while ensuring that this alteration would not adversely impact the rights and well-being of the participants, as outlined in \cite{common_rule_2018}, \S 46.116(e)(2).

Whenever individuals were actively engaged with the robot and interactions were recorded, we sought their consent post-interaction. Additionally, we requested permission to utilize any images or footage in which they appeared. Consent was documented through recorded verbal assent, following the guidelines in \cite{common_rule_2018} \S 46.117(c)(1). We opted for verbal assent because obtaining signed consent would be the sole record connecting the subjects to the research, and the primary risk involved was potential harm due to a breach of confidentiality, with the research itself posing minimal risk.

In line with the conventions of field research conducted in public spaces, we did not request consent from passersby, who were only incidentally involved in the study, despite the potential argument that employing concealed "wizard" operators could be considered a form of deception. \citet{sommers2013forgoing} provide a more in-depth ethical discussion on such studies. Additionally, we acquired written permission from the business improvement district responsible for managing the study location and a certificate of insurance to cover any inadvertent damages that might result from the robot's deployment.

\subsection{Interviews}
We conducted interviews with all individuals who have had any form of interaction with the robots, whether that interaction was explicit or implicit. The field interviewer should demonstrate a keen awareness of subtle interactions. It is evident that those who directly engage with the robots should be included in the interviews, as they are active participants in the interaction. Conversely, those with negative sentiments towards the robots may subtly avoid interacting with them. Such nuanced interactions should not be overlooked, as ignoring them will bias the interviewed population.

Unlike lab environments, there is no pause in public deployments. The robots continuously interact with people one after another. Given the volume of interactions in public spaces, we recommend having at least two field interviewers interviewing people concurrently. 

Our protocol dictated that the process of obtaining consent and conducting interviews should take place after the interaction has concluded. We advise researchers to wait until there is a definite end to the interaction activities, such as when the robots leave or the participants disperse, before initiating the interview process. This is to ensure that the interaction is not disrupted. Additionally, to prevent influencing the interactions of other participants, researchers should refrain from immediately approaching participants after their interaction. The only exception to this is if participants are exiting the study area, to minimize the risk of revealing the researcher's connection with the robots to bystanders. Typically, there is a sequence of robot interactions followed by interviewing various individuals present, albeit separately. However, this approach does have some limitations. Despite our efforts to minimize our presence in the area separate from the robots, there is still a possibility that those in the vicinity may associate the researcher with the robots, which could potentially impact the authenticity of the experience we strive to create.

\subsection{Emergent Protocol}
Field studies often have an exploratory element, leading to situations where pre-designed study protocols might not be fully applicable in the actual field setting. Unforeseen circumstances during deployment can pose challenges in following established protocols. In cases where protocol adjustments do not impact the data-gathering process, we advise seeking guidance from the University's IRB department and making appropriate amendments to the protocol. Otherwise, it may be best to redesign the study. 

Early on during the deployment, we realized that due to uneven surfaces, the robots tended to get stuck from time to time. Since we always had two robots deployed in the field at the same time, it was natural that the other robot came to "push" the stuck robot to maintain the narrative of them being "autonomous." It was not until both robots got stuck or one robot lost connection that the wizards got another field member's attention to intervene. While these procedures were not documented in the protocol before the study, they emerged naturally in the field from practices, and we added them to the protocol for later deployments.

Other changes in protocols can be triggered by unexpected edge cases. Rarely, the wizards were involved in a serendipitous conversation around the robots with bystanders unaware of the wizards' identities. In such scenarios, it was difficult for wizards to hold a conversation since they could not lie to the bystanders (e.g. "We don't know if the robots are autonomous", "we don't know who owns them", etc.), which will make the study a deception study. On the other hand, revealing their identity proactively will break the illusion. The IRB protocol must provide clear guidelines for handling such situations based on the study context. In our case, after a discussion with the University IRB office, we decided to tell the bystanders that the robots were part of a university's study and that we would like to answer any further questions offline.

\section{Guidelines Regarding the Robots}

\subsection{Hardware}
As we deploy robots in the field, we prioritize making the setup portable. In general, there are two sets of hardware for robots' field deployments: the robots themselves and the communication infrastructure.

Unlike in-lab studies, field deployments require higher standards for mobile robots' durability and robustness. It also calls for higher movement speed, stronger torque, and more power. To avoid reinventing the wheels, we repurposed motors and chassis from old hoverboards to provide the power and durability a robot needs to survive in the wild. On top of the original hoverboard metal chassis, we mounted a fiberglass-reinforced dolly to hold the trash barrel on top. There is always a tradeoff between robot portability and computation. For Wizard-of-Oz deployments, the computation (planning and control) is handled by wizards (researchers), which allows us to utilize smaller single-board computers (SBC). Specifically, a Raspberry Pi 4 serves as the brain for each robot, and the motors are managed by \href{https://odriverobotics.com/shop/odrive-v36}{ODrive v3.6}.

The core of an in-the-field system is a reliable communication network, which is crucial for data transfer and teleoperation. In our study, we used WiFi to establish the connection since it is relatively easy to set up and provides a reasonable coverage range. Specifically, we used a \href{https://www.netgear.com/de/home/mobile-wifi/hotspots/mr1100/}{Netgear Nighthawk M1 mobile router}  to host a WiFi signal and a \href{https://www.amazon.com/dp/B07RN9PFNX?ref=ppx_yo2ov_dt_b_product_details&th=1}{Netgear Orbi WiFi Range Extender} to increase the covered area. 

\begin{figure}[ht!]
    \centering
    \includegraphics[width=0.45\textwidth]{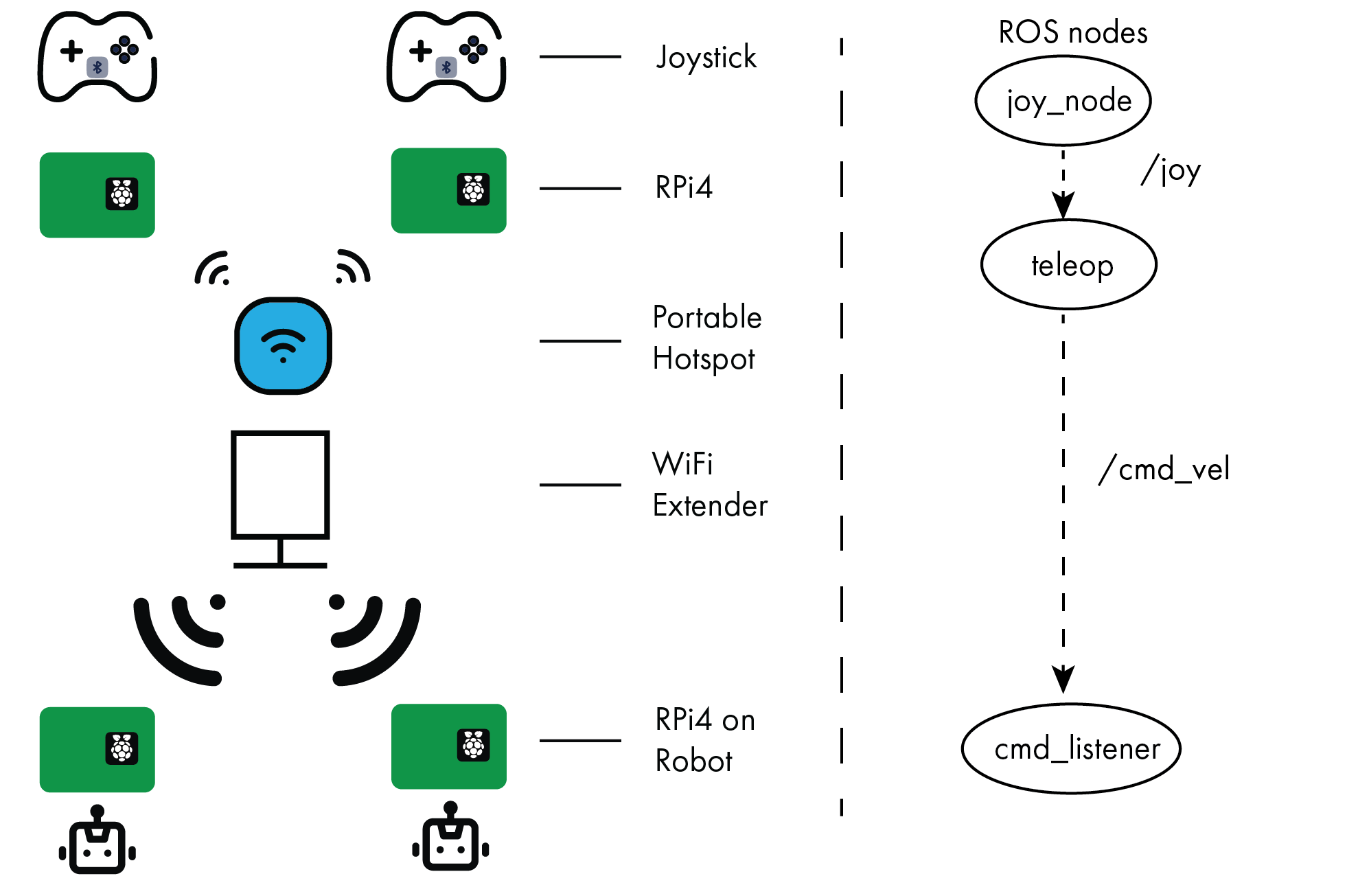}
    \caption{Left: hardware communication setup. Right: ROS structure. Joysticks are connected to RPi4s via Bluetooth. Wizards' commands are sent to the RPi4s on the robot via Wifi. On the ROS side, the process \texttt{joy\_node} reads raw signals from joysticks and publishes them on the \texttt{/joy} topic. The \texttt{teleop} node converts these raw signals to twist commands, which are published to \texttt{/cmd\_vel} topic.}
    \Description{This figure is divided into left and right sections with a dashed line. On the left is a systematic diagram showing the hardware architecture of our deployment in a top-down diagram. On the top are two joysticks, which are connected to two RPi 4 via Bluetooth. The two RPi 4s are connected to the RPi 4 on the robots via WiFi. On the right is the systematic diagram showing the ROS software architecture in a graph. The joy_node connects with the teleop node through the /joy topic. The teleop node connects to the cmd_listener node through the /cmd_vel topic.}
    \label{hardware_sys}
\end{figure}

The wizard controls the robot through commercial gamepads. Since the robots are deployed in a Wizard-of-Oz way, we found the smallest \href{https://a.co/d/96gOwq6}{gamepad} possible on the market. (However, we later noticed that there is a tradeoff between size and comfort as wizards complained that tiny buttons hurt their fingers.) The choice of gamepad largely depends on the type of control signals wizards need to send. If all control commands can be expressed in binary, a gamepad with only buttons will suffice (and those gamepads can be even smaller). However, for mobile robots where control signals are in the continuous space, joysticks or pressure-sensitive analog triggers are required. 
Overall, we recommend using as many off-the-shelf parts as possible to increase robustness. Our system diagram is shown in Figure~\ref{hardware_sys}.

\subsection{Software}
This section offers actionable guidance for researchers conducting pilot studies with robots in field settings. We focus on robots that are teleoperated on-site without automation, using software primarily for controlling the robots and collecting data from various sensor streams.

Given the complexity of field operations, maintaining software modularity is crucial for ease of system startup and troubleshooting on-site. Our software is developed using the Robot Operating System (ROS1) framework. There are multiple advantages to using ROS. Primarily, it is a well-established platform in the robotics community and boasts a robust ecosystem. Additionally, ROS's computational graph design facilitates process separation into nodes, enhancing system modularity. In our setup, we employ three ROS nodes: one to establish a Bluetooth connection with the joystick, another to translate joystick signals into robot control commands, and a third to relay these commands to the robot's hardware. Notably, only the last node operates directly on the robot.

Despite ROS being a preferred framework for robotics development, its field application can be challenging. Managing multiple nodes often involves operating several terminals. To streamline this, we suggest utilizing ROS launch files and creating shell scripts that initiate ROS automatically upon device boot-up. The current ROS user interface is not particularly user-friendly for field deployments, but improvements are anticipated in the future.

\subsection{On Behavior Design}
The action space of the robots depends on both hardware and software. For mobile robots in populated spaces designed to interact with humans, we prefer differential drive robots over Ackerman drive robots due to their high maneuverability~\cite{mekhtarian2014mechanically}. Differential drive robots can rotate in place, providing more possible "expressions" than Ackerman drive robots. For example, the robot can turn left and right at a relatively high frequency to "wiggle."

Since there are usually multiple buttons on a joystick, it is a good practice to pre-program an action sequence for each button. For instance, we programmed the button "B" to make the robot wiggle. We suggest that pre-programmed sequences should be short, modular, and interruptable.

A standard joystick returns two float numbers within the range of -1 to 1, one represents forward and backward, and the other represents left and right. Theoretically, one could control a differential drive robot with just one joystick. However, in practice, we realized that it is easier to "express" when separating linear and angular motion control commands and mapping them to two different joysticks on the same gamepad. For example, in our deployment, the left joystick controls linear velocity only, and the right joystick controls angular velocity. This allows wizards to control robots more precisely than using a single joystick controller.

\subsection{Emergency Button and Watchdog}
It is common practice that every robot must come with an emergency stop button for safety concerns, even in lab settings. It is more crucial for robots to have an emergency switch when deployed in public spaces. However, since robots deployed in public spaces are usually mobile, it is counter-intuitive to have a physical button. For our study, we took the approach where the motors are in an idle state by default. The wizards need to hold a button to operate the robot intentionally. Releasing the button will stop the robots' motions right away. 

In situations where the robots are out of signal coverage or signal delay, it is necessary to consider error handling from both software and hardware levels. In the field, robotic operations are commonly about reactiveness. Depending on the application, it may be beneficial to discard old, unprocessed commands and prioritize new command messages. Carefully choosing the queue size of ROS subscribers that listen to the command messages may help handle system behavior when communication is delayed. 

However, software alone is not enough. In our early days of testing, we noticed that the lower-level controller (on ODrive) latched on to the last received message and kept the execution running (for example, the robot went out of range right before receiving the STOP command). When this happens, there is nothing one can do from the software perspective, the system is not communicating. This is when a watchdog implemented on the motor controller board itself comes in handy. During operations, the system should keep sending control signals to feed the watchdog (even if the command is 0 velocity). A watchdog should kick in when no signals are received within a predefined threshold (1 second in our case, usually due to lost connections) to shut down the motors regardless of previous commands. For convenience, the researcher should have a handy script to clear the watchdog's lock and reset the robots.

\subsection{Testing in Lab vs. in the Field}
While testing in lab settings is necessary, piloting at deployment locations is also crucial~\cite{mitka2012safety}. 
Prior to deployment, we envisioned and prepared for a multitude of potential failures, yet we discovered this was insufficient. The challenge extended to basic functionalities; steering the robot in a straight line through urban streets proved arduous. This experience imparts a crucial lesson: Pursuing perfection in laboratory conditions can be counter-effective. It is essential, albeit daunting, to venture into field testing well before achieving a fully operational prototype. Embrace the simplicity of cardboard models and subject them to the rigors of real-world conditions. This proactive approach can unveil invaluable insights and practical challenges that laboratory environments simply cannot replicate.

Some mistakes are hard to fix once you have a fully operating system. For example, we decided to mount the hoverboard chassis in the perfect center of the dolly so that the robot could spin in place perfectly. However, in practice, the front and back caster wheels kept lifting the motorized wheels in the center when they ran into small rocks, and the robot would just get stuck. Moving the wheels off the center requires redesigning the arrangement of electronics inside the robot and re-drilling holes in the metal chassis and the dolly. Thus, do not hesitate to go to the field with your cardboard duck-taped prototypes.

\subsection{Debugging in the Field}
\textit{Try not to debug in the field.} It always happens, but researchers should try to eliminate the chances. Before going to the fields, researchers should understand their system well, being aware of all the weak points in both hardware and software. 

Hardware should be as robust as possible. Tighten all screws, and use threadlocker (e.g., \href{https://www.henkel-adhesives.com/de/de/\%C3\%BCber-uns/unsere-marken/loctite.html}{Loctite}) to secure them (they \textit{will} come off otherwise). Use heat shrinks to cover all the wire connections, and zip tie wires to the chassis to avoid dangling. There is usually no good way to fix a hardware failure onsite. 

On the software side, restarting is always your best friend. Bring a portable monitor if possible, as \texttt{ssh} is not usually the most efficient way to debug, especially when the problem is network-related. When working with multiple robots that are physically identical, beyond labeling the robots themselves, label the joysticks they paired with as well. When a robot fails during deployment, walking to the robot and reading its label will break the overall interaction experience for users, especially for wizard-of-oz studies. Instead, it is best practice to figure out the robot's identity from its paired joystick and resolve the issue remotely. 

\section{Guidelines on Study Site}
\subsection{Permission}
Getting permission to run the study in public took a substantial amount of time. Depending on the ownership of the location, the difficulty varies. Overall, in \anon{NYC}, we found that plazas managed by local business improvement districts (BIDs) are most approachable and supportive. During our deployment, both BIDs have onsite security during the day. \anon{Downtown Brooklyn Management} even provided onsite storage and connected us with their field teams for any last-minute requests. Locations managed by public facilities, such as museums and libraries, are also relatively easy to access with the caveat that the population demographics are less diverse. Parks are usually owned by the government, which has its own application procedure for research-related activities. The review process can be long, and the allowed duration can be restricted. 

\subsection{On Deployment Locations}
Selecting a deployment location involves coordination with various stakeholders, such as city officials, management bodies, and users of the space. From a broader perspective, factors like local demographics, population density, and weather conditions are important considerations. More specific details, such as the physical layout of the area, its operational hours, and the patterns of pedestrian movement, also require careful attention.

Ideally, a deployment site should be an open area with ample room for navigation and interaction. Yet, it's equally important to observe and respect how people typically use the space, adhering to any unspoken social norms. For instance, in our case, there were complaints about robots posing a risk to people after we temporarily changed our deployment site to a triangle traffic island. Despite the study area being an open plaza, the presence of a subway entrance at its center and multiple pharmacies nearby influenced how people moved through the space. The triangular shape of the plaza and adjacent shops created certain implicit social norms governing pedestrian traffic. The introduction of robots disrupted these norms, leading to numerous complaints.

In our studies, we deploy the robots at two plazas: one touristy and one local. We notice that the neighborhood of the study location may also influence the deployments. For example, people at the local plaza are more likely to have an attachment to the location due to their personal history and provide more contextual feedback. For example, we interviewed a truck driver who has lived around the area for 10 years. He reflected on the changes around the plaza he had witnessed in the decade. It is also more likely to have reoccurring interactions with the same people since hanging out at the local plaza is part of their daily routines. On the other hand, the touristy plaza benefits from its large foot traffic, where hundreds of people from different cultural backgrounds visit the plaza with different intentions. Therefore, if the focus of the research is to stress test robotic systems with a diverse group, we recommend touristy plazas. If the focus of the research is to investigate how people adapt to the introduction of robotics systems in their daily routine,  we believe a local plaza would be a better fit.

\subsection{Instrumenting the Space}
Cameras mounted in the environment can provide a unique perspective on how deployed robots transformed the public space, which is unintuitive to observe from cameras mounted on the robot. In contrast to controlled laboratory settings where researchers can easily install sensors like cameras, mounting sensors in public spaces poses significant challenges. Concealing sensors is often the preferred approach to prevent any form of priming effect, as well as to avoid triggering individuals who may wish to avoid being recorded (in countries where recording in public is legal).

\section{Guidelines on Data Collection}
The rule is always to record as much data as possible that your hardware enables. In public robot deployments, there are three primary sources of data to consider: the robots, the environment, and the operators (often referred to as 'wizards'). Capturing video and audio from the robots' point of view is essential, as it offers a firsthand perspective of each interaction. Additionally, collecting data on aspects like the robots' speed is crucial for understanding their behavior. Documenting the environment is also key, as it helps in analyzing how the space changes over time. Lastly, in scenarios where multiple robots are managed by several wizards, the interactions between these wizards can yield valuable insights for developing multi-agent cooperation systems in the given contexts.

One common challenge in data collection in the wild is noise. Temporary Visual obstacles, such as market umbrellas, and audio noise, such as urban background sound, should be considered before actual data collection. If possible, when interviewing participants or recording wizards' conversations, researchers should place the camera closer to the subjects or provide a microphone. Post-processing data collected in the field is very difficult. 

In our deployments, we mounted 360-degree cameras to record robot-centric data. We mounted two GoPro cameras on the exterior of a coffee shop located at the southeast corner of the plaza to record the entire plaza. We also installed a camera in front of the wizards to record their activities as well.

\section{Guidelines on Press and Media Coverage}
   The public nature of our deployment is poised to garner attention from sources like local journalists and the media. While such publicity is advantageous for showcasing our research and aligns with the interests of collaborating organizations, it can be less beneficial for studies conducted in public spaces over a long duration. Notably, in our second deployment, which occurred a year after the first and in a different area, passersby still recognized the robots from our earlier study. To address this, we controlled media exposure during the deployment. We arranged for media coverage on the last study day when the research was complete, and we could accommodate additional attention.

\section{Conclusion}
Our experiences from field deployments have taught us that robots in public spaces require different considerations and priorities compared to in-lab studies. Through sharing our experiences and lessons learned, we aim to inspire HRI community to conduct more studies outside laboratory settings. Additionally, we advocate for a collaborative effort, urging researchers who have conducted beyond-lab robot deployments to share their methodologies via a common GitHub repository. We believe this joint endeavor will advance the HRI field, making it more relevant and influential in shaping the future of interactive digital societies.

\begin{acks}
We express our gratitude to the Village Alliance and Downtown Brooklyn Management for permitting us to conduct the study at Astor Place and Albee Square, respectively. This research received research funding from Tata Consulting Services and Urban Tech Hub at Cornell Tech. This work was also supported by a fellowship of the German Academic Exchange Service (DAAD). We extend our appreciation to the wizards who played integral roles behind the scenes: Rei Lee, Saki Suzuki, Ricardo Gonzalez, Stacey Li, Maria Teresa Parreira, and Jorge Pardo Gaytan.
\end{acks}

\bibliographystyle{ACM-Reference-Format}
\bibliography{robots_in_the_field}
\end{document}